\documentclass{article}

\usepackage{microtype}
\usepackage{graphicx}
\usepackage{subfigure}
\usepackage{booktabs} % for professional tables
\usepackage{amsmath}
\usepackage{amssymb}

\usepackage{hyperref}

\DeclareMathOperator{\softplus}{softplus}
\DeclareMathOperator{\supremum}{sup}
\newcommand{\eq}[1]{\begin{equation} #1 \end{equation}}
\newcommand{\al}[1]{\begin{align} #1 \end{align}}
\newcommand{\pa}[1]{\left( #1 \right)}

\renewcommand{\P}{\mathbb{P}}
\newcommand{\E}{\mathbb{E}}

\newcommand{\calL}{\mathcal{L}}

\newcommand{\calD}{\mathcal{D}}

\renewcommand{\Pr}{\mathbb{P}_\text{r}}
\newcommand{\Pg}{\mathbb{P}_\text{g}}
\newcommand{\Prhat}{\hat{\mathbb{P}}_\text{r}}
\newcommand{\Pghat}{\hat{\mathbb{P}}_\text{g}}

\usepackage[accepted]{icmlw2019generalization}

\icmltitlerunning{Investigating Under and Overfitting in GANs}

\begin{document}

\twocolumn[
\icmltitle{Investigating Under and Overfitting in\\Wasserstein Generative Adversarial Networks}



\icmlsetsymbol{equal}{*}

\begin{icmlauthorlist}
\icmlauthor{Ben Adlam}{goo,ai}
\icmlauthor{Charles Weill}{goo}
\icmlauthor{Amol Kapoor}{goo}
\end{icmlauthorlist}

\icmlaffiliation{goo}{Google Research, New York, NY, USA}
\icmlaffiliation{ai}{Work done as a member of the Google AI Residency program (g.co/brainresidency).}

\icmlcorrespondingauthor{Ben Adlam}{adlam@google.com}

\icmlkeywords{Machine Learning, ICML}

\vskip 0.3in
]



\printAffiliationsAndNotice{}  % leave blank if no need to mention equal contribution

\begin{abstract}
We investigate under and overfitting in Generative Adversarial Networks (GANs), using discriminators unseen by the generator to measure generalization. We find that the model capacity of the discriminator has a significant effect on the generator's model quality, and that the generator's poor performance coincides with the discriminator underfitting. Contrary to our expectations, we find that generators with large model capacities relative to the discriminator do not show evidence of overfitting on CIFAR10, CIFAR100, and CelebA.
\end{abstract}

\section{Introduction}
\label{introduction}

Generative adversarial networks (GANs) are a widely used type of generative model that have found success in many data modalities \cite{goodfellow2016nips}. For image datasets, GANs have been able to generate diverse, high-fidelity samples that are almost indistinguishable from real images \cite{brock2018large, karras2017progressive, karras2018style}. However, achieving such results remains difficult due to many types of training failure \cite{salimans2016improved}, a lack of effective measurements of model quality \cite{theis2015note, barratt2018note}, and the need for substantial hyperparameter tuning \cite{lucic2018gans, kurach2018gan}.

There is a growing body of work centered on investigating the statistical issues faced by GANs \cite{arora2017generalization, arora2017gans, bai2018approximability} and optimization challenges specific to GANs \cite{mescheder2017numerics, nagarajan2017gradient,  hsieh2018finding, rafique2018non}. We add to this work by analyzing the effect of function class complexity on the model quality of the generator and GAN generalization. We begin by discussing the theoretical underpinnings of under and overfitting for GANs. We then introduce the \emph{auxiliary discriminator} and \emph{independent discriminator} as mechanisms to help measure GAN performance. We use these tools to empirically probe how model complexity impacts GAN outcomes.

\section{Preliminaries}

In generative modeling, the goal is to find model parameters $\theta_g$ that minimize a divergence $\mathcal{D}$ between the true data distribution $\Pr$ and the model distribution $\Pg$. These divergences can be specified as the solution to a variational problem. For example, the Kantorovich-Rubinstein duality states Wasserstein distance is
\eq{\label{eq: krd}
    \calD_\text{W}\pa{\Pr, \Pg} = \supremum_{f\in L_1} \calL_f(\Pr, \Pg),
}
where the supremum is over all 1-Lipschitz functions and $ \calL_f(\Pr, \Pg) := \E_{X\sim\Pr} f(X) -  \E_{Y\sim\Pg} f(Y)$. In general, these divergences cannot be easily estimated from samples (let alone optimized with respect to) unless the distributions $\Pg$ and $\Pr$ have a specific parametric form. 

In practice, optimizing over all 1-Lipschitz functions is infeasible, so GANs replace $L_1$ with a function class of neural nets $F$ that are constrained to be 1-Lipschitz by weight clipping \cite{arjovsky2017wasserstein}, gradient penalization \cite{gulrajani2017improved}, or spectral normalization \cite{miyato2018spectral}. This leads to the notion of neural net divergence $\calD_\text{NN}$ \cite{arora2017generalization}. The learning task of the discriminator $f$ is to find parameters $\theta_f$ that maximize $\calL_f(\Pg,\Pr)$.

Despite relaxing $L_1$ to $F$ solving for $\calD_\text{NN}$ is still difficult, since the expectations in $\calL_f(\Pr, \Pg)$ cannot be computed exactly and the parameter space of $F$ is non-convex. We can at least compute the expectations for a training set $(\Pghat, \Prhat)$ and optimize $\theta_f$ with respect to this to obtain some $\tilde{f}$, and then consistently estimate $\calL_{\tilde{f}}(\Pg,\Pr)$ using a test set. Note even if $\tilde{f}$ approximately achieves the supremum on the training set and has no generalization gap to the test set, we cannot conclude that $\tilde{f}$ is approximately $\calD_\text{NN}(\Pg,\Pr)$, only that it provides a lower bound. All of this conspires to make using $\calD_\text{NN}$ as an objective measure of the generator's model quality challenging.

\subsection{Under and Overfitting}

For any metric, a model's generalization gap is the difference between the metric's value on the true data distribution less its value on the training set. As a model changes, classical machine learning theory divides the behavior of the generalization gap into two regimes: underfitting and overfitting. Overfitting is when the metric is improved on the training set at the expense of its value on the true data distribution. In supervised learning, metrics such as accuracy can be easily measured on the training set and estimated on the true data distribution using an independent test set, but as we have discussed estimating divergences is more difficult.

Generative models can also suffer from under and overfitting. Specifically, for GANs, we say a \emph{generator $g$ is overfitting} if it is minimizing $\calD(\Pghat, \Prhat)$ at the expense of increasing $\calD(\Pg, \Pr)$---that is, the generalization gap between $\calD(\Pghat, \Prhat)$ and $\calD(\Pg, \Pr)$ increases faster than $\calD(\Pghat, \Prhat)$ decreases. We say a \emph{discriminator $f$ is overfitting} if it is increasing $\calL_f(\Pghat, \Prhat)$ at the expensive of decreasing $\calL_f(\Pg, \Pr)$. This can happen when $F$ is too complex or the training set size is too small, since standard Rademacher complexity arguments can be used to show $\calL_f(\Pghat, \Prhat)$ is close to $\calL_f(\Pg, \Pr)$ for all $f$, and hence the supremums must also be close. See \cite{arora2017generalization} for an example where this generalization fails because $F$ is too complex. 

Underfitting is harder to define, but the comparison of $\calD_\text{NN}(\Pg, \Pr)$ to $\calD_\text{W}(\Pg, \Pr)$ is relevant, as we motivated the former as a tractable version of the later. Specifically, if $F$ has insufficient capacity $\calD_\text{W}$ may be significantly larger, and we argue this corresponds to underfitting.

\subsection{Additional discriminators}

To compare to the original discriminator $f_\text{O}$ (that the generator learns from), we train two additional discriminators that from the perspective of NN divergence are solving the same variational problem \eqref{eq: krd}. The first $f_\text{A}$, called an \emph{auxiliary discriminator}, has a different, random initialization, learns with the same loss, but does not provide gradients to the generator. The idea being that its optimization is similar to the original discriminator, (it has the same training data, it is learning a non-stationary objective). The second $f_\text{I}$, called an \emph{independent discriminator}, attempts to abstract away many of the details of GAN training and simply compute NN divergence. Using either the training set $D_\text{train}^1$, which the generator learned from, or $D_\text{train}^2$, which was not used during training of the GAN, and an equal number of samples from an already trained generator, the independent discriminator optimizes $L_D$. Moreover, we can either use the same architecture for all three discriminators, or we can vary the original discriminator's architecture while fixing the auxiliary and independent discriminators to act as \emph{baselines}.

Since they all solve \eqref{eq: krd}, the divergences computed by these discriminators, $\calL_{f_\text{O}}$, $\calL_{f_\text{A}}$, and $\calL_{f_\text{I}}$ can be compared. The auxiliary discriminator provides an additional measure of the model quality of the generator, and by comparing it to the original discriminator we can detect underfitting when $\calL_{f_\text{O}} < \calL_{f_\text{A}}$, as $\calD_\text{W}$ is greater than $ \calL_{f_\text{A}}$

The divergence $\calL_{f_\text{I}}$ can be used in the same ways as $\calL_{f_\text{A}}$ outlined above, but it can also be used to probe for overfitting in the generator. We use the difference between $\calL_{f_\text{I}}(\hat{\P}_{D_\text{train}^2}, \hat{\P}_g)$ and $\calL_{f_\text{I}}(\hat{\P}_{D_\text{train}^1}, \hat{\P}_g)$ to approximate the generator's generalization gap and a large gap could suggest overfitting.

\section{Experimental Setup}

We consider the image datasets CIFAR10, CIFAR100, and CelebA at the resolutions $32\times32\times3$, $32\times32\times3$, and $64\times64\times3$ respectively. We split the training data in half, denoted $D_\text{train}^1$ and $D_\text{train}^2$, where $| D_\text{train}^i| = 25K$ for CIFAR10 and CIFAR100, and $| D_\text{train}^i| = 45K$ for CelebA. Additionally, we use a test set, denoted $D_\text{test}$ where $| D_\text{test}| = 10K$ to monitor overfitting in the discriminators. 

We train each Wasserstein GAN for 500K steps to reach an equilibrium and use a batch size of 64. As per \cite{miyato2018spectral}, optimization is done with Adam using a learning rate of 0.0001, $\beta_1$ of 0.5, and $\beta_2$ of 0.999. The discriminator and generator are both updated once in each training step. Contrastingly, it sufficed to optimize the independent discriminator using SGD with cosine decay over 100K steps to reach convergence. 

We use the losses
\al{\label{loss}
	\nonumber L_D &= \softplus\pa{\sum_i D(G(z_i)) - \sum_{i}D(x_i)} \\
	\text{and } L_G &= \softplus\pa{-\sum_i D(G(z_i)) }
}
for the discriminator and generator respectively, where $D$ and $G$ are the functions computed by the GAN and the sum is over a batch of real data $x_i$ and noise $z_i$. This softplus loss can be thought of as a smooth version of hinge loss and a monotonic function of \eqref{loss} \cite{miyato2018spectral}. Since we interpret the discriminators as computing a divergence between distributions, we plot the negation of $L_D$ without the softplus in all figures.

\begin{figure}[ht]
\vskip 0.2in
\begin{center}
\centerline{\includegraphics[width=0.85\columnwidth]{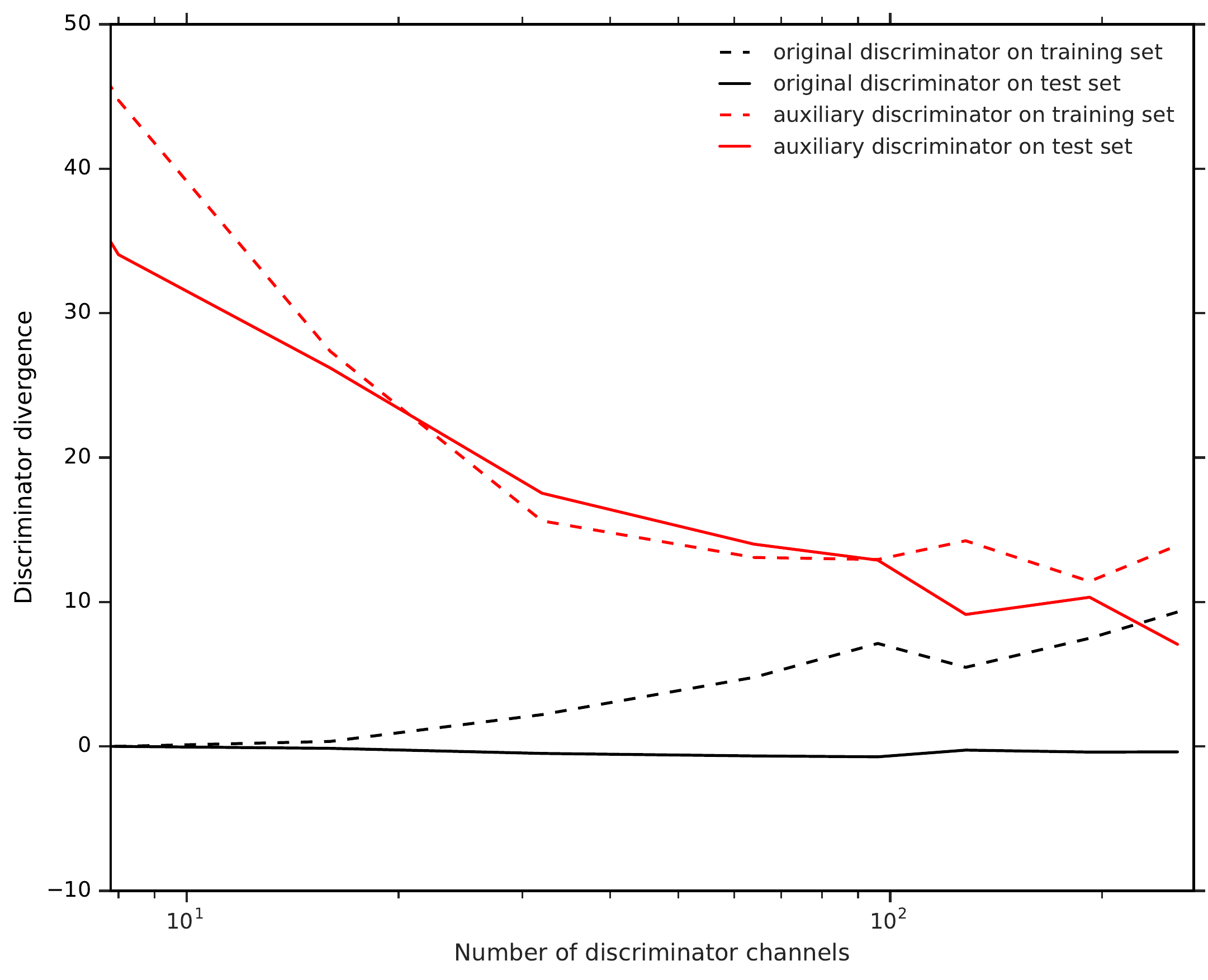}}
\caption{Results for CIFAR10. We observe little generalization gap (difference between solid and dashed lines) for the auxiliary discriminator, whereas the original discriminator's gap appears to increase with model capacity. Note it completely fails to generalize to the test set. The decreasing auxiliary divergence suggest that the model quality of the generator improves with additional discriminator model capacity, but this trend is not seen in the original discriminator--in fact, it is reversed on the training set. The significant gap between original and auxiliary divergences is symptomatic of underfitting, but interestingly since the original and auxiliary discriminators have the same architecture, we find the cause must be an interaction between model capacity and optimization.}
\label{f1}
\end{center}
\vskip -0.2in
\end{figure}

We use a DCGAN as in \cite{miyato2018spectral}. The generator has batch normalization, no spectral normalization, and ingests samples from a standard 128 dimensional Gaussian. The generator's architecture is fixed through all experiments. The discriminator has spectral normalization (to enforce the Lipschitz continuity) and no batch normalization. The number of channels in the discriminator's convolutional layers are changed during experiments to vary model capacity from 8 to 256.

\begin{figure}
    \centering
    \begin{minipage}{.6in}%
        \centerline{\includegraphics[width=.8in]{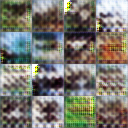}}
    \end{minipage}
    \quad\,
    \begin{minipage}{.6in}%
        \centerline{\includegraphics[width=.8in]{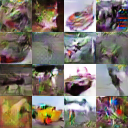}}
    \end{minipage}
    \quad\,
    \begin{minipage}{.6in}%
        \centerline{\includegraphics[width=.8in]{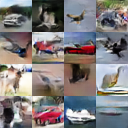}}
    \end{minipage}
    \quad\,
    \begin{minipage}{.6in}%
        \centerline{\includegraphics[width=.8in]{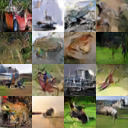}}
    \end{minipage}
    \\
    \begin{minipage}{.6in}%
        \centerline{\includegraphics[width=.8in]{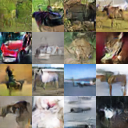}}
    \end{minipage}
    \quad\,
    \begin{minipage}{.6in}%
        \centerline{\includegraphics[width=.8in]{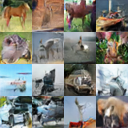}}
    \end{minipage}
    \quad\,
    \begin{minipage}{.6in}%
        \centerline{\includegraphics[width=.8in]{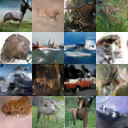}}
    \end{minipage}
    \quad\,
    \begin{minipage}{.6in}%
        \centerline{\includegraphics[width=.8in]{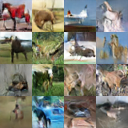}}
    \end{minipage}
    \caption{Samples from generators trained on CIFAR10 ordered in increasing number of discriminator channels. Qualitatively the generator's image quality seems to improve as the discriminator's model capacity increases.}%
    \label{images}
\end{figure}

We also compare the divergence computed by the discriminators and Fr\'{e}chet Inception distance (FID), which is a commonly used metric for evaluating GANs \cite{salimans2016improved, heusel2017gans}. We compute FID using 10K samples for both $D_\text{test}$ and for $D_\text{train}^1$.

\section{Results}

To begin with we discuss under and overfitting in the discriminators. In general, we observed no overfitting in any of the discriminators, and mostly the gap between the discriminator's divergence on the training and test sets was small but sometimes noisy. The only consistent generalization gap was found in the original discriminator's divergence on $D_\text{train}^1$ and $D_\text{test}$. The discriminator's divergence on $D_\text{train}^1$ decreased consistently with the number of channels, whereas its divergence on $D_\text{test}$ continued to fluctuate around zero, which implies no overfitting. Note that a divergence of zero is what is achieved by a random discriminator. The same behavior was observed on CIFAR100 (see supplementary figures). For CelebA (see supplementary figures), the test divergence decreased with number of channels, which caused the generalization gap to stay relatively constant. It is puzzling that the generator is able to learn from and produce a better model using a discriminator that completely fails to generalize. We did however see a significant difference between the original discriminator and the auxiliary and independent discriminators (see Figure~\ref{f1}) that suggested underfitting in the original discriminator.

For CIFAR10 and CIFAR100, the divergences computed by the discriminators for each generator were broadly similar, regardless of the discriminator's architecture. On CelebA, the discriminators agreed in the rank ordering of divergences, but the independent discriminators often reported larger divergences than the auxiliary discriminators.

\begin{figure}[ht]
\vskip 0.2in
\begin{center}
\centerline{\includegraphics[width=0.85\columnwidth]{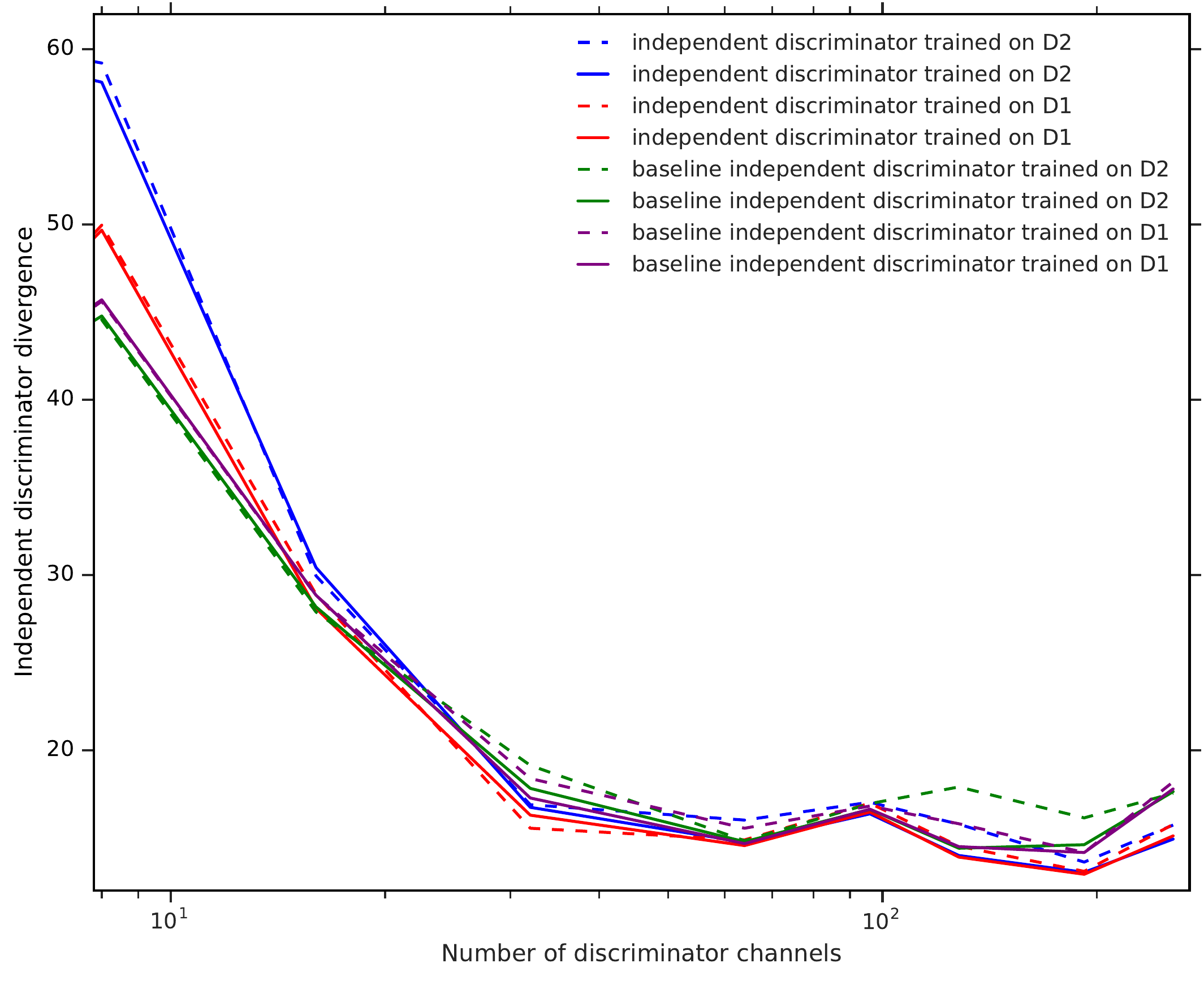}}
\caption{The independent discriminators generalize well for CIFAR10. See difference between test (solid) and training (dashed) sets. Their divergences are broadly similar whether the architecture matches the original discriminator or is fixed at a baseline of 64 channels. We note that for 8 channels, $\calL_{f_\text{I}}(\hat{\P}_{D_\text{train}^2}, \hat{\P}_g)$ was larger than $\calL_{f_\text{I}}(\hat{\P}_{D_\text{train}^1}, \hat{\P}_g)$, which would suggest a generalization gap for the generator, but this not replicated in the baseline 64 channel independent discriminators.}
\label{f2}
\end{center}
\vskip -0.2in
\end{figure}

\begin{figure}[ht]
\vskip 0.2in
\begin{center}
\centerline{\includegraphics[width=0.85\columnwidth]{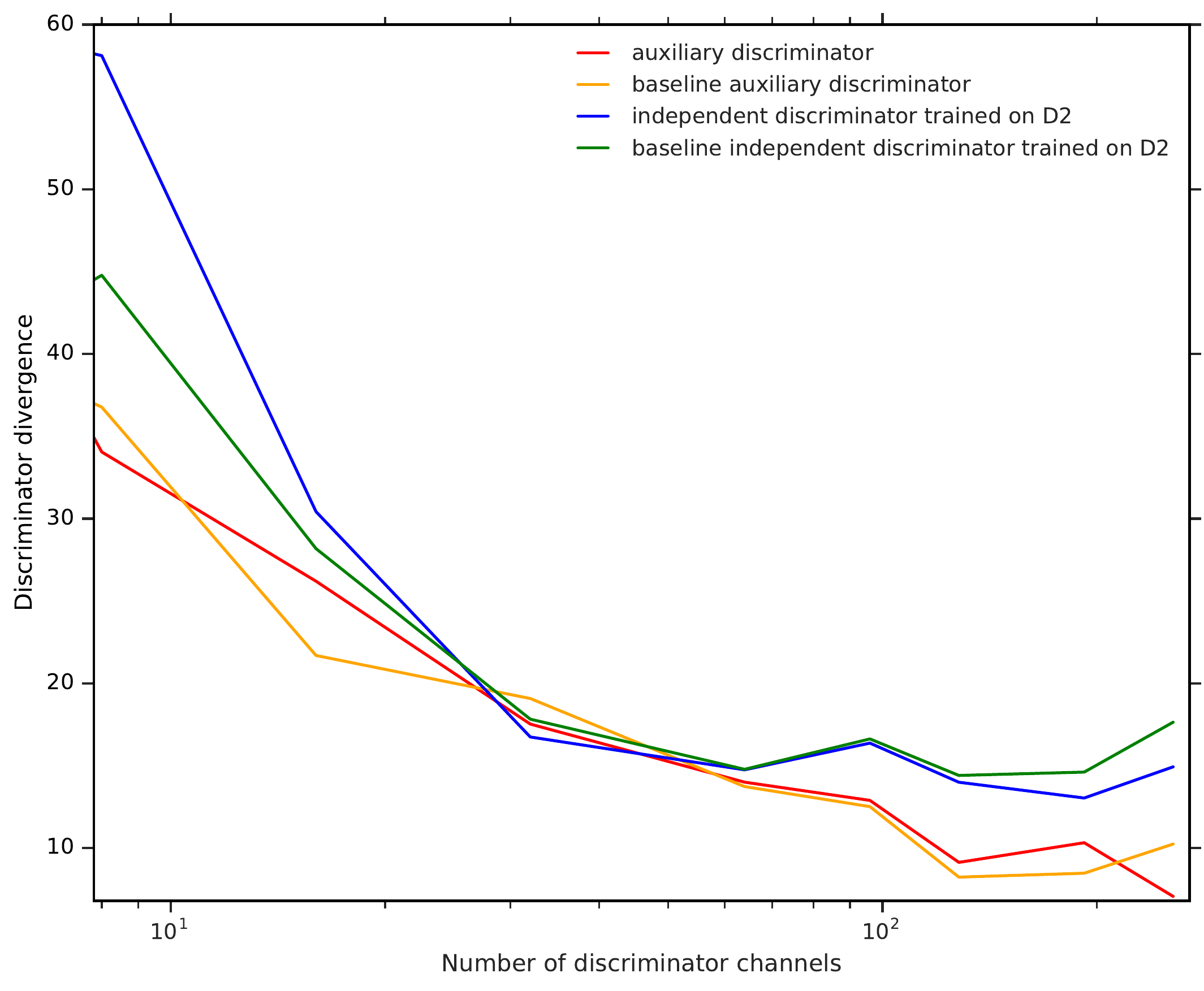}}
\caption{For CIFAR 10, the auxiliary and independent discriminators achieve broadly similar divergences despite differences in how they are optimized, but at the extremes the auxiliary discriminators' divergences are smaller.}
\label{f3}
\end{center}
\vskip -0.2in
\end{figure}

Turning to the generator, we note the divergences computed by the auxiliary and independent discriminators generally decreased with number of channels---indicating that the generator is actually producing a better model. Note that this was true whether the architectures used for the auxiliary and independent discriminators matched that of the original discriminator, or if they had a consistent, baseline of 64 channels.

For the generator, we observed no overfitting when evaluated by the independent discriminator or by FID. This is somewhat intuitive given the samples in Figure~\ref{images}. Taking a different perspective, the gap between the original and auxiliary discriminators, could be viewed as a type of overfitting by the generator in the following sense: when the discriminator has few channels and the generator is relatively overparameterized, it lowers its loss, as computed by the original discriminator, to zero without decreasing the auxiliary discriminator's divergence. This is similar to a classifier decreasing its loss, at the expense of accuracy. Interestingly, the difference between the original and auxiliary discriminator reduces as the original discriminator's generalization gap increases.

\begin{figure}[ht]
\vskip 0.2in
\begin{center}
\centerline{\includegraphics[width=0.85\columnwidth]{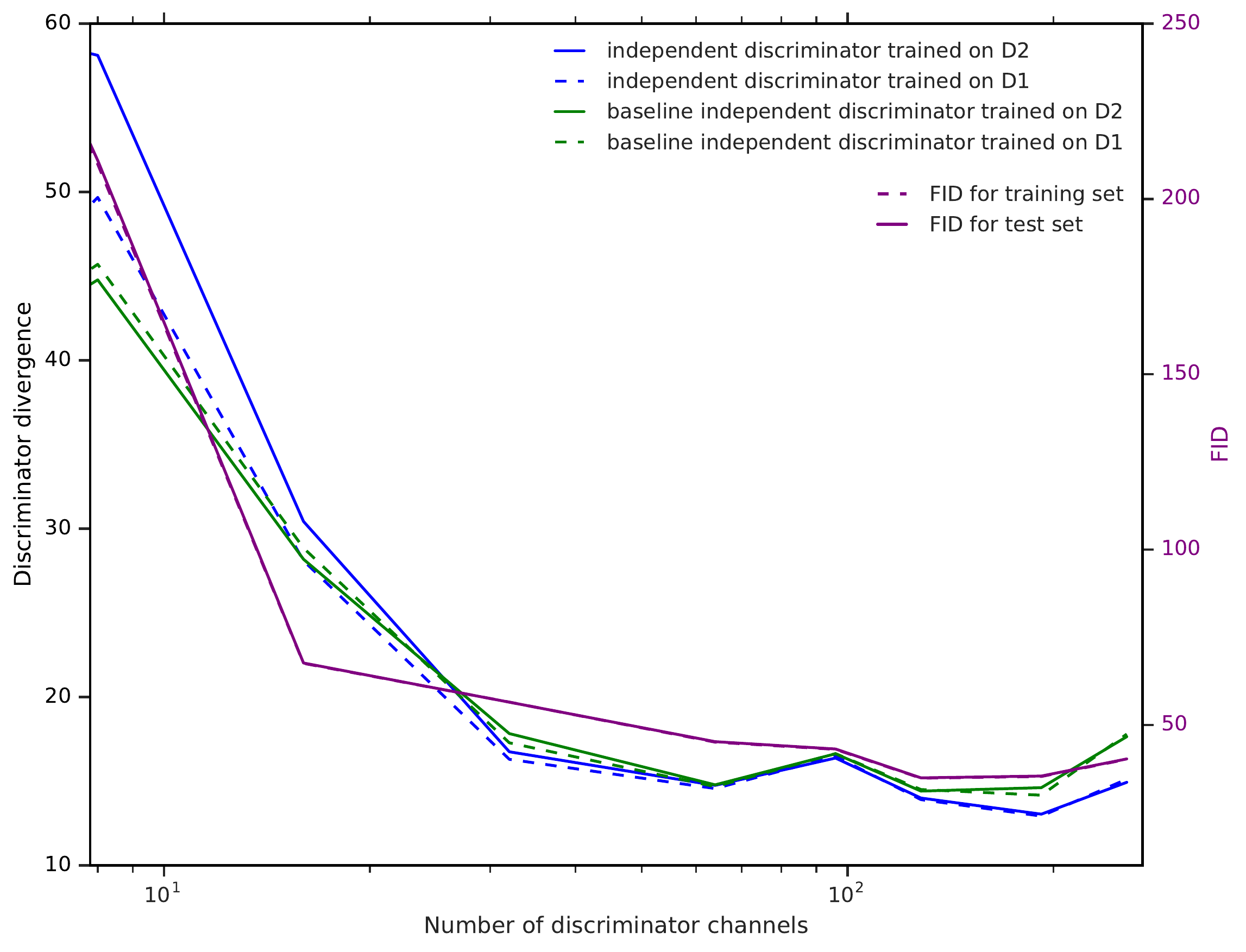}}
\caption{FID correlates relatively well with the independent discriminator's divergence on CIFAR10. There is little difference between the independent discriminator's divergence or FID on the training set and test set, indicating that the generator is not overfitting with respect to this metric.}
\label{f4}
\end{center}
\vskip -0.2in
\end{figure}

\section{Conclusion and further research}

We find that the relative model capacity of the discriminator has a significant effect on the model quality of the generator. However, we do not observe any overfitting in either the generator or discriminator.

We saw that the original discriminator either does not generalize at all or has a large generalization gap. Moreover, its divergence does not correlate with the generator's model quality, whereas the auxiliary and independent discriminators divergences do. These divergences also correlate with FID. We wonder whether these divergences might be used to evaluate GANs for data modalities where a canonical trained embedding, like Inception, is not available.

In the future, we plan to extend this analysis to investigating the effects of dataset size and batch size on GANs. While we have focused here on Wasserstein GANs, we hope to extend out analysis to other types of GAN.

\bibliography{paper}
\bibliographystyle{icml2019}

\appendix

\section{Additional figure for CIFAR10}
\begin{figure}[ht!]
\vskip 0.2in
\begin{center}
\centerline{\includegraphics[width=0.85\columnwidth]{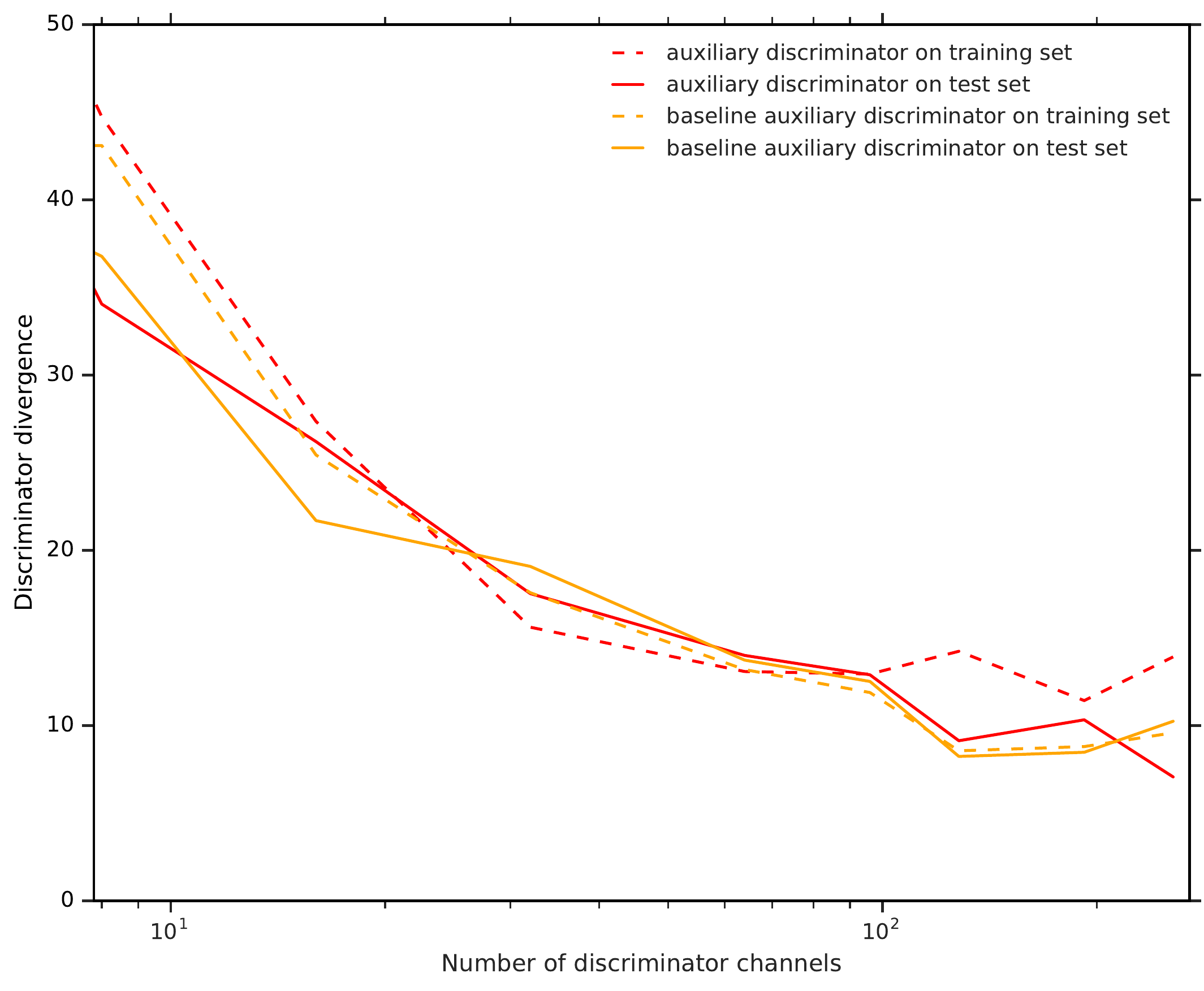}}
\end{center}
\vskip -0.2in
\end{figure}

\section{Figures for CIFAR100}

\begin{figure}[ht!]
\vskip 0.2in
\begin{center}
\centerline{\includegraphics[width=0.85\columnwidth]{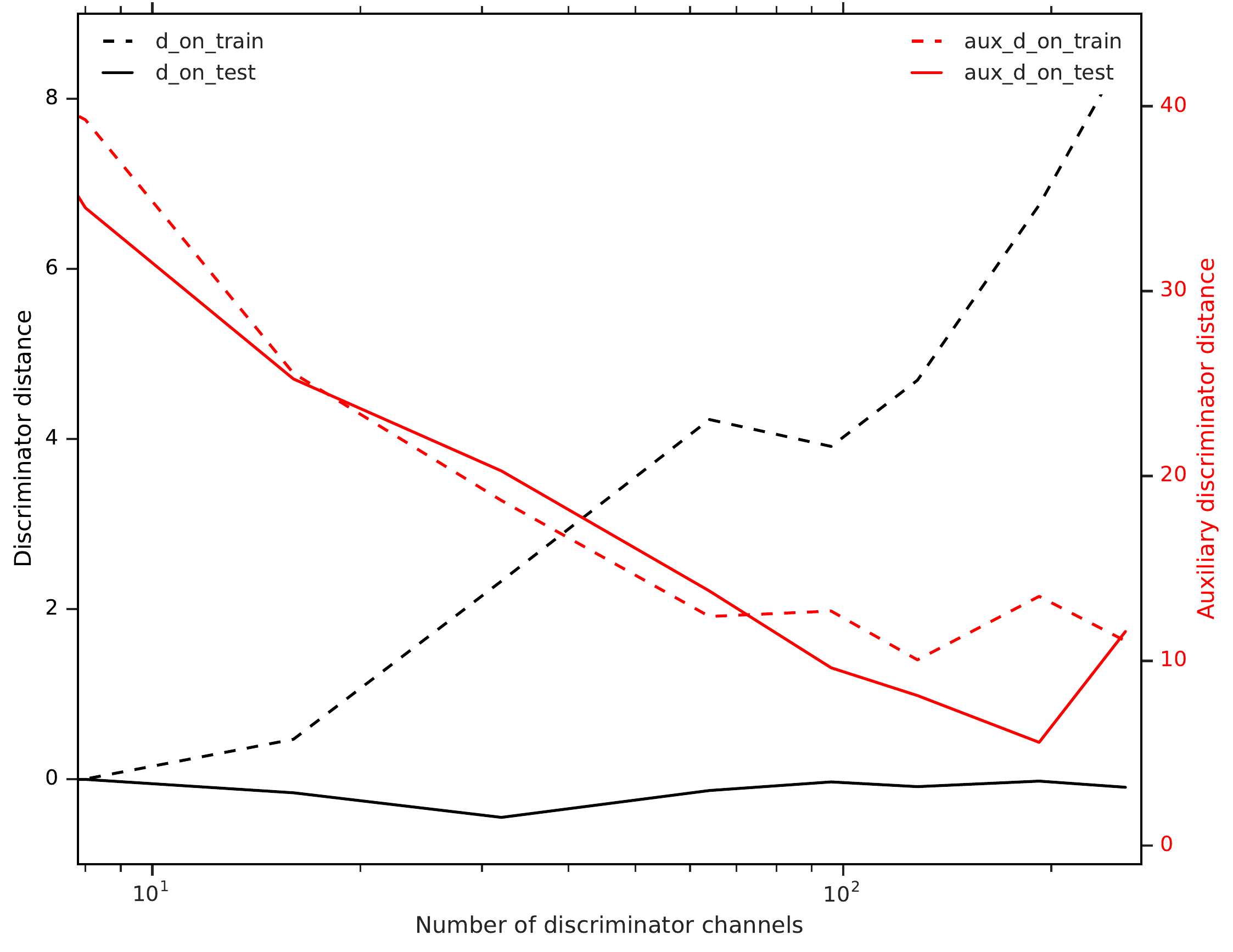}}
\end{center}
\vskip -0.2in
\end{figure}

\begin{figure}[ht!]
\vskip 0.2in
\begin{center}
\centerline{\includegraphics[width=0.85\columnwidth]{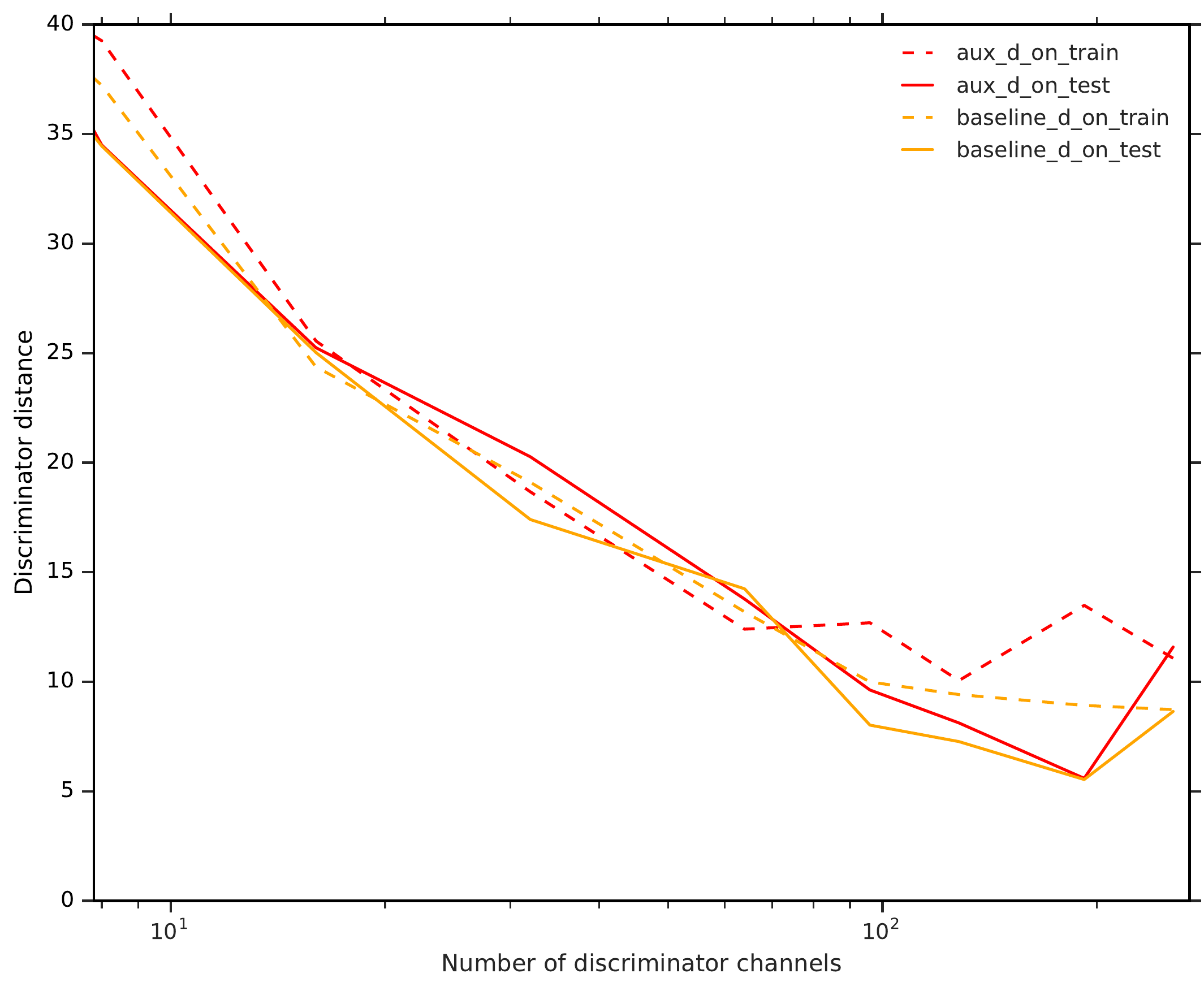}}
\end{center}
\vskip -0.2in
\end{figure}

\begin{figure}[ht!]
\vskip 0.2in
\begin{center}
\centerline{\includegraphics[width=0.85\columnwidth]{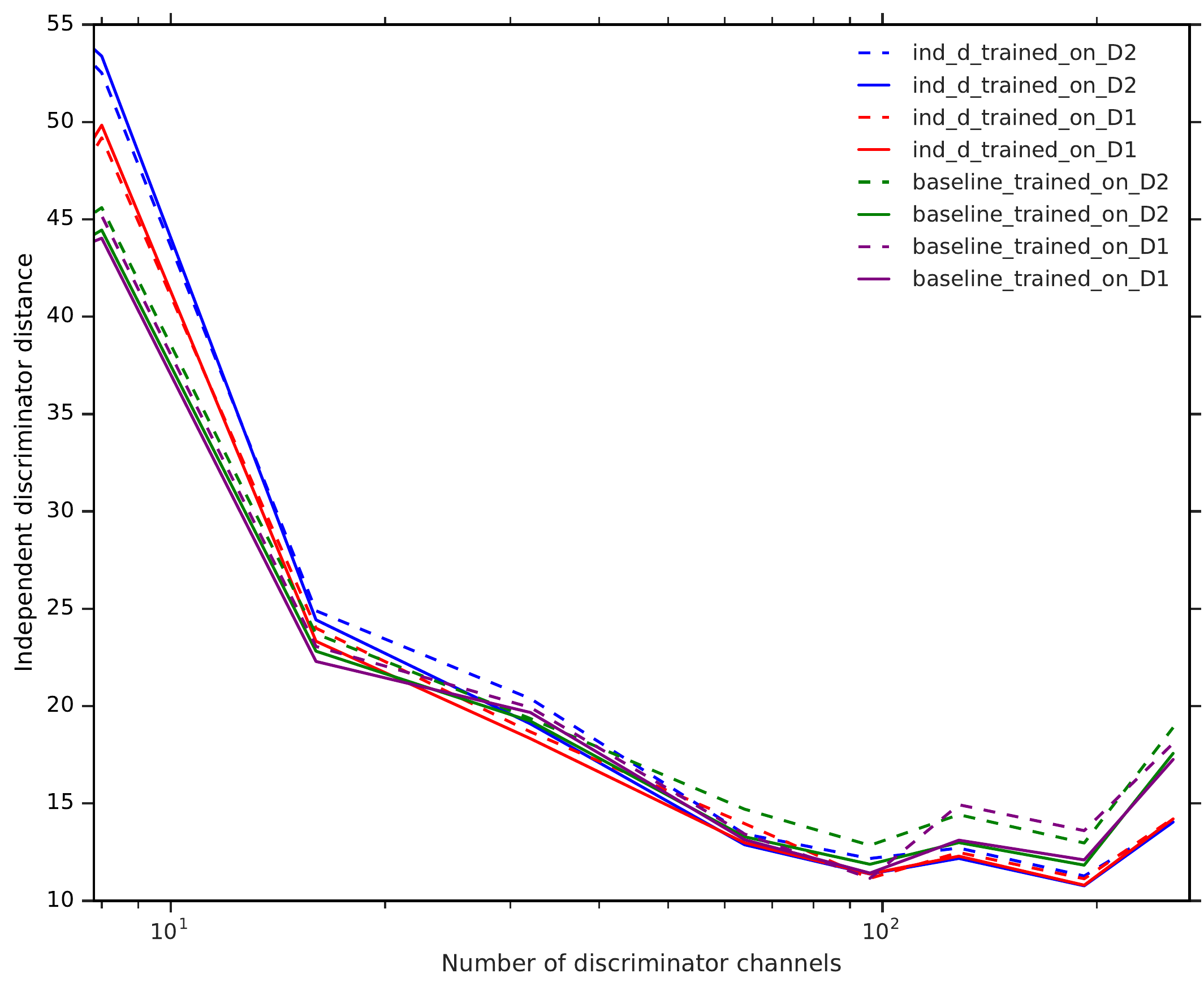}}
\end{center}
\vskip -0.2in
\end{figure}

\begin{figure}[ht!]
\vskip 0.2in
\begin{center}
\centerline{\includegraphics[width=0.85\columnwidth]{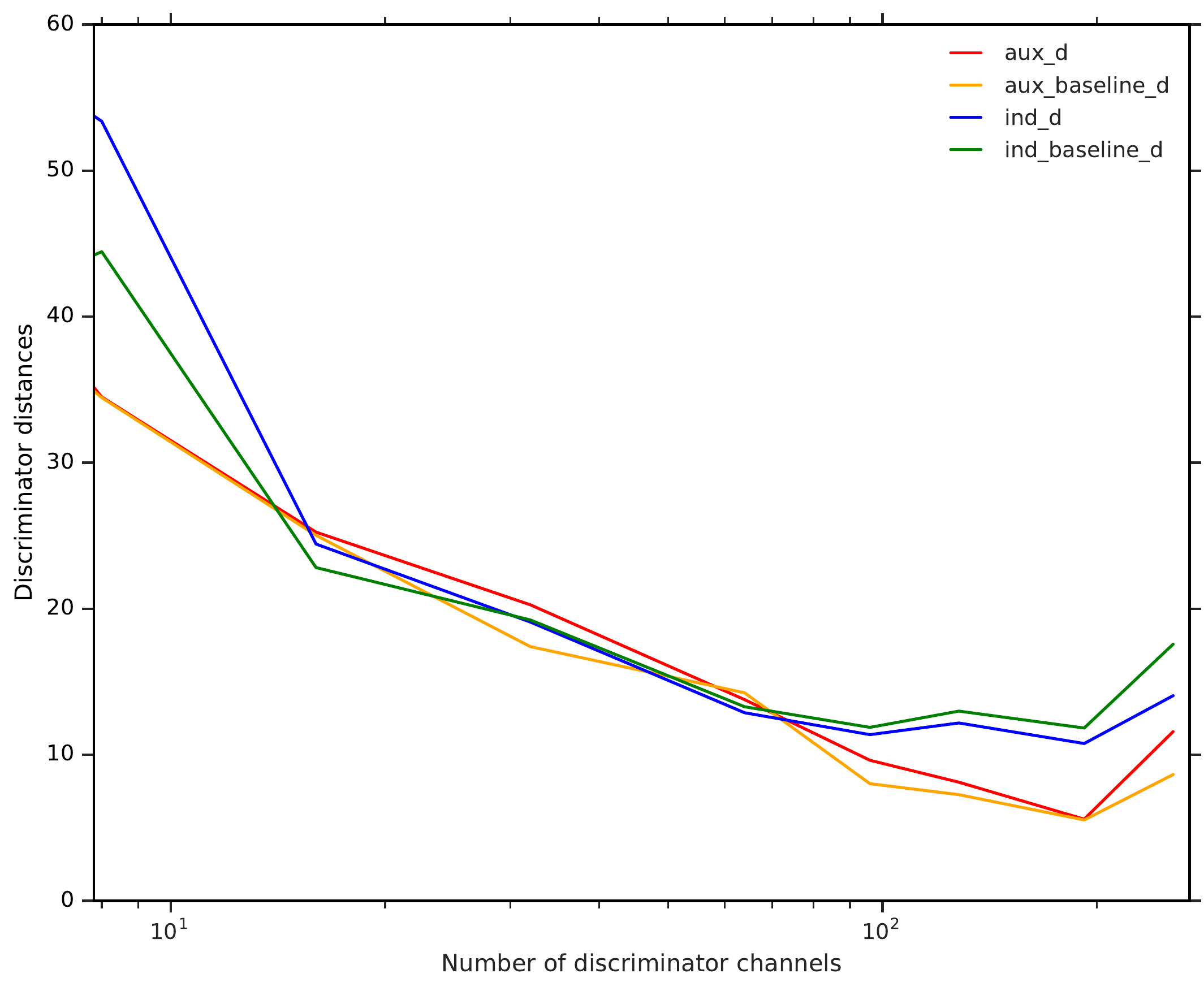}}
\end{center}
\vskip -0.2in
\end{figure}

\begin{figure}[ht!]
\vskip 0.2in
\begin{center}
\centerline{\includegraphics[width=0.85\columnwidth]{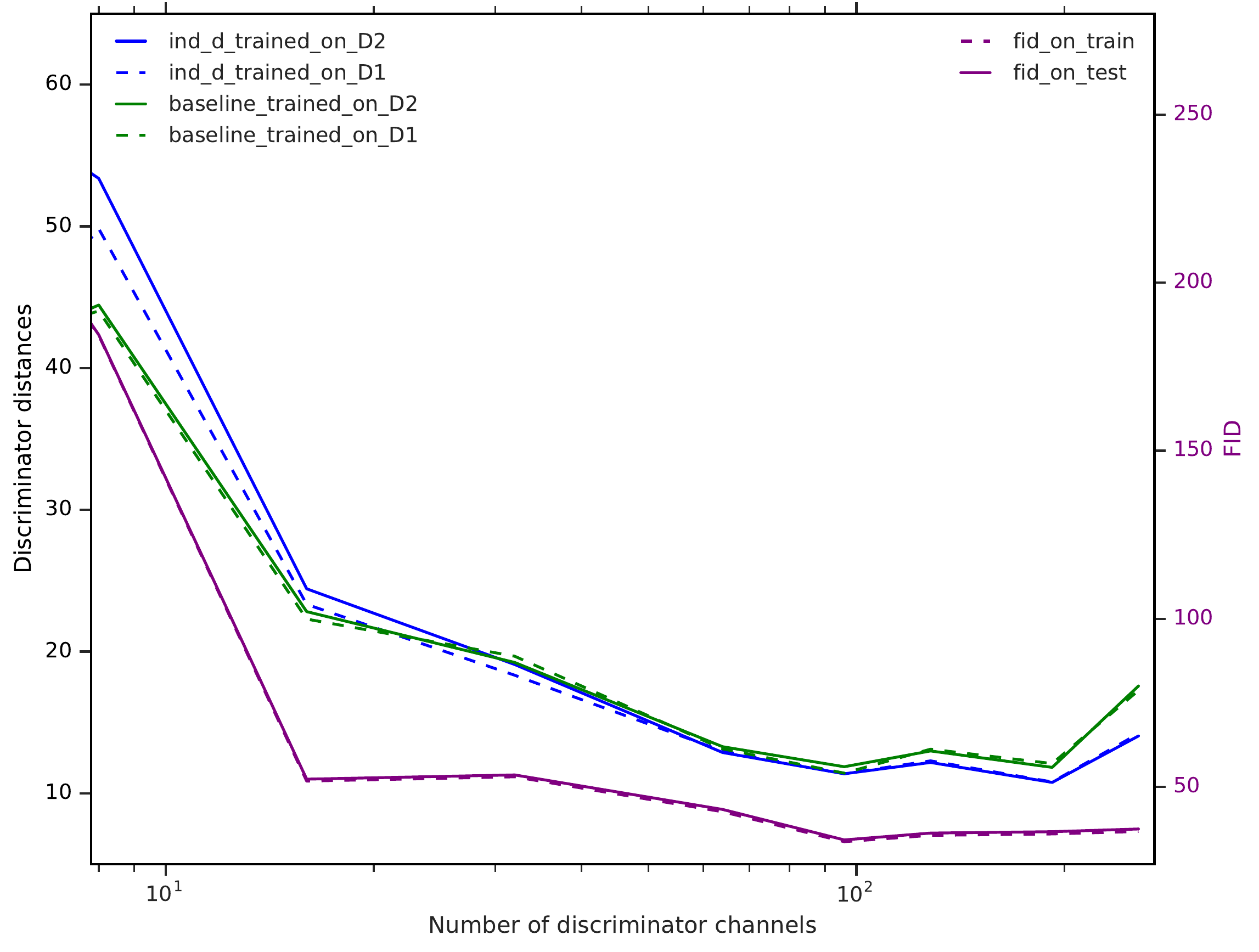}}
\end{center}
\vskip -0.2in
\end{figure}
\newpage
~
\newpage

\section{Figures for CelebA}

\begin{figure}[ht!]
\vskip 0.2in
\begin{center}
\centerline{\includegraphics[width=0.85\columnwidth]{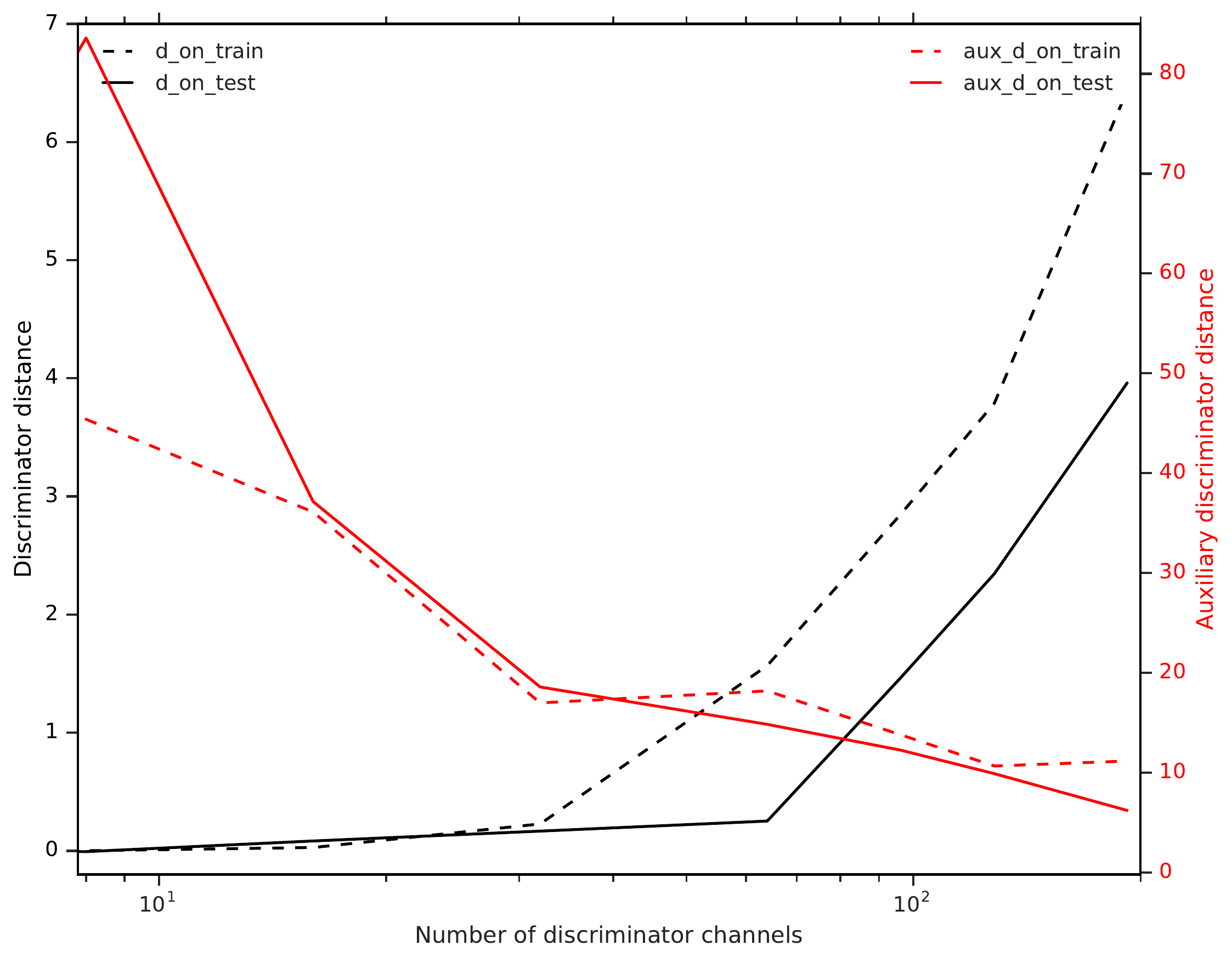}}
\end{center}
\vskip -0.2in
\end{figure}

\begin{figure}[ht!]
\vskip 0.2in
\begin{center}
\centerline{\includegraphics[width=0.85\columnwidth]{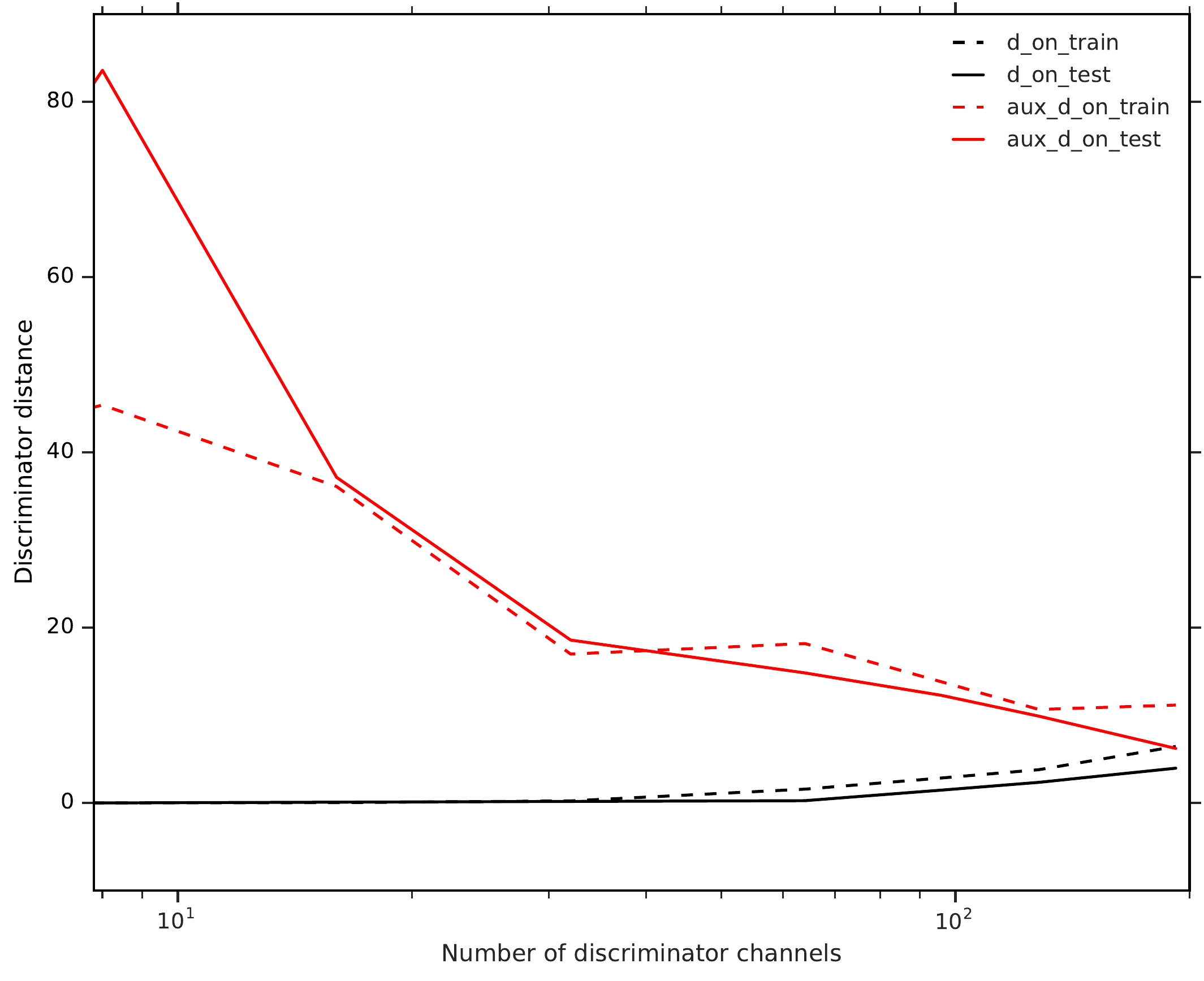}}
\end{center}
\vskip -0.2in
\end{figure}

\begin{figure}[ht!]
\vskip 0.2in
\begin{center}
\centerline{\includegraphics[width=0.85\columnwidth]{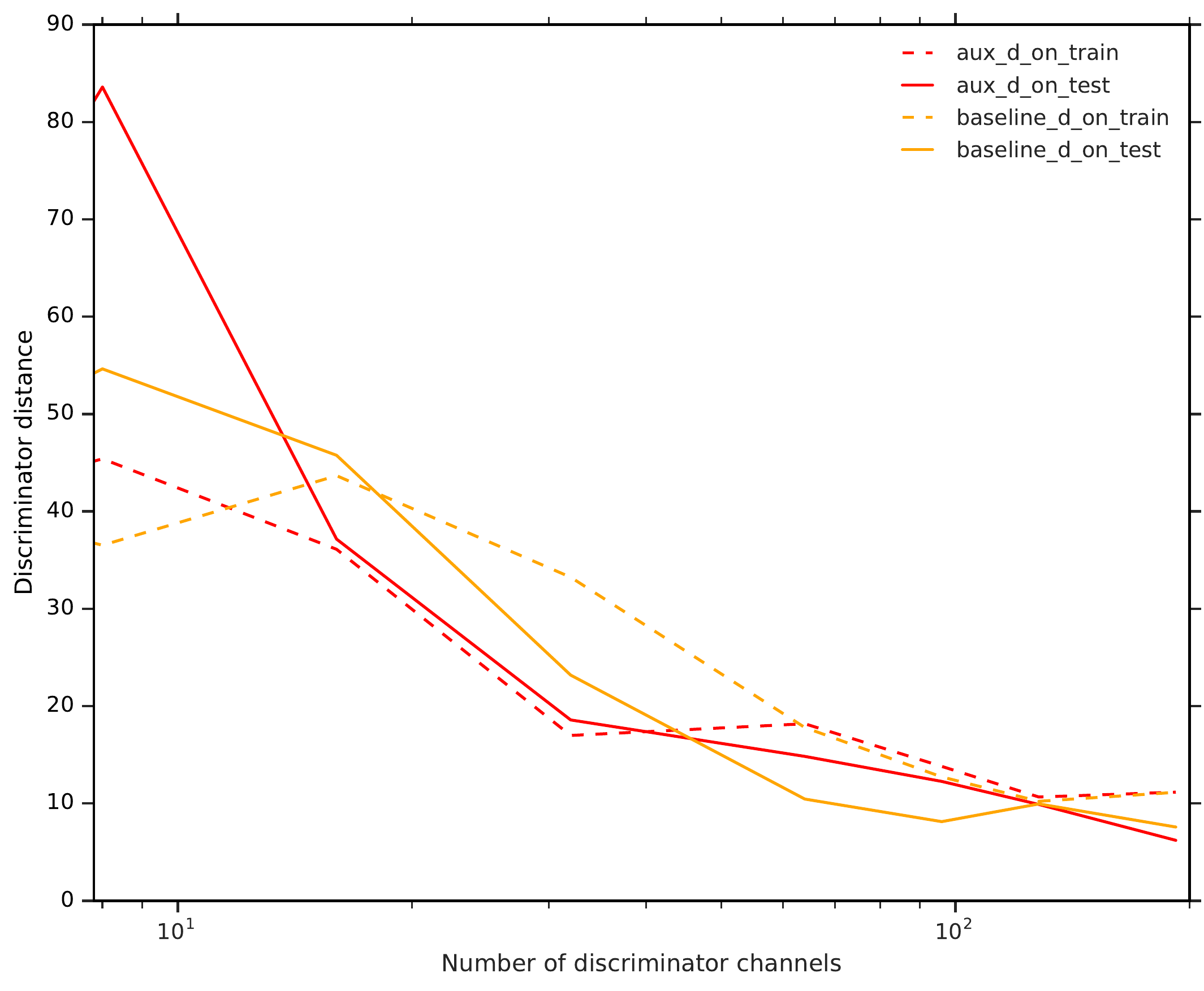}}
\end{center}
\vskip -0.2in
\end{figure}

\begin{figure}[ht!]
\vskip 0.2in
\begin{center}
\centerline{\includegraphics[width=0.85\columnwidth]{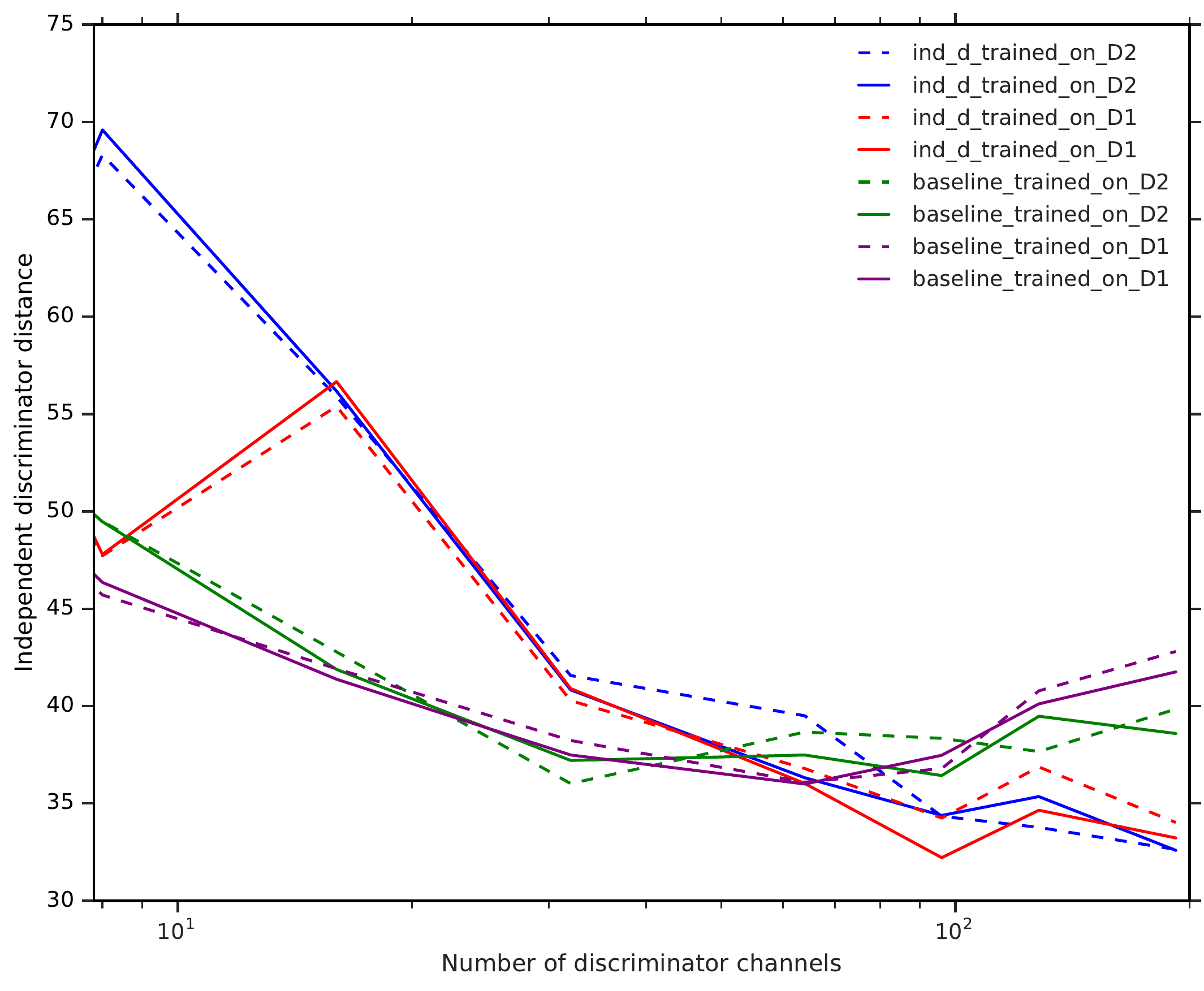}}
\end{center}
\vskip -0.2in
\end{figure}

\begin{figure}[ht!]
\vskip 0.2in
\begin{center}
\centerline{\includegraphics[width=0.85\columnwidth]{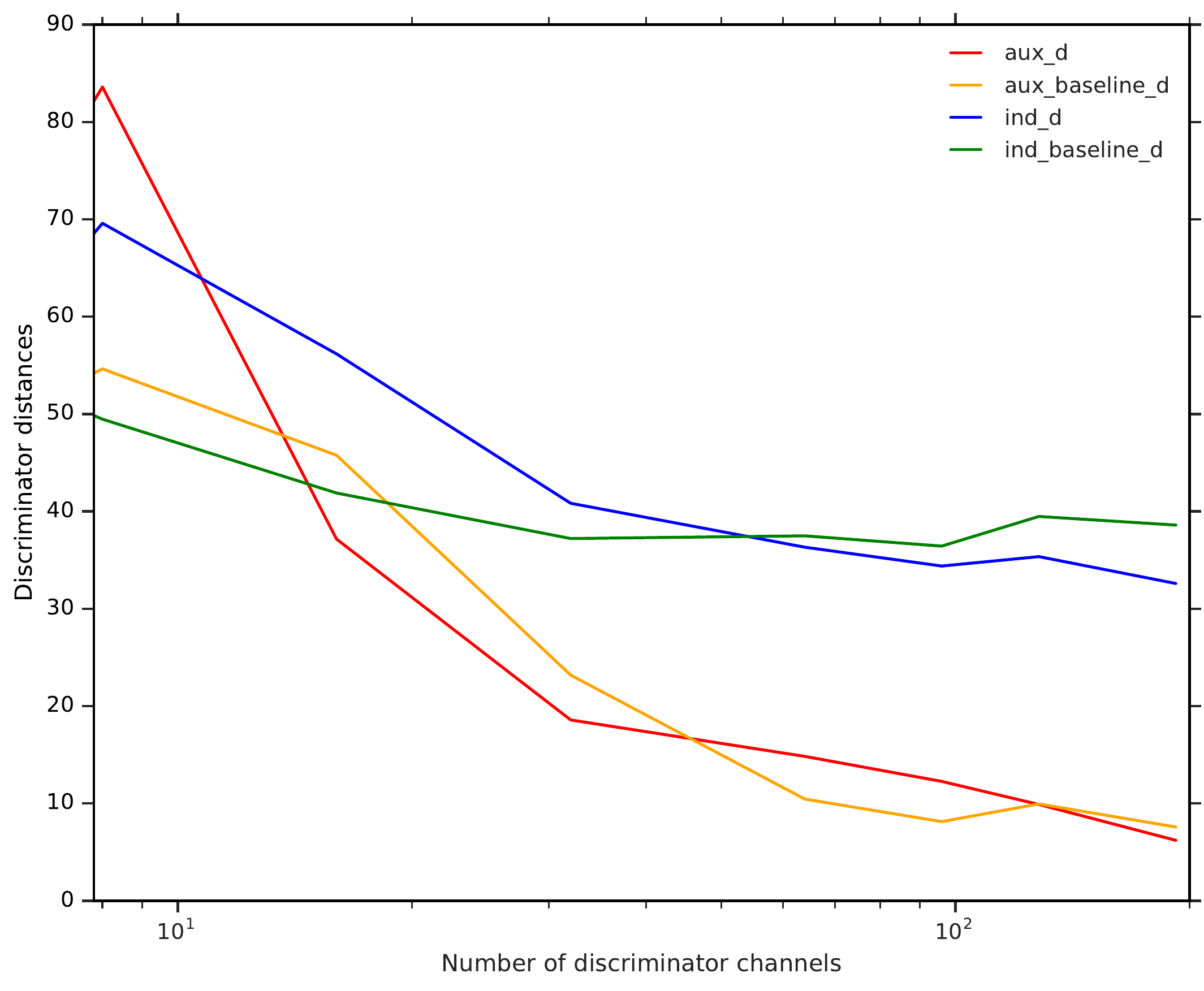}}
\end{center}
\vskip -0.2in
\end{figure}

\begin{figure}[ht!]
\vskip 0.2in
\begin{center}
\centerline{\includegraphics[width=0.85\columnwidth]{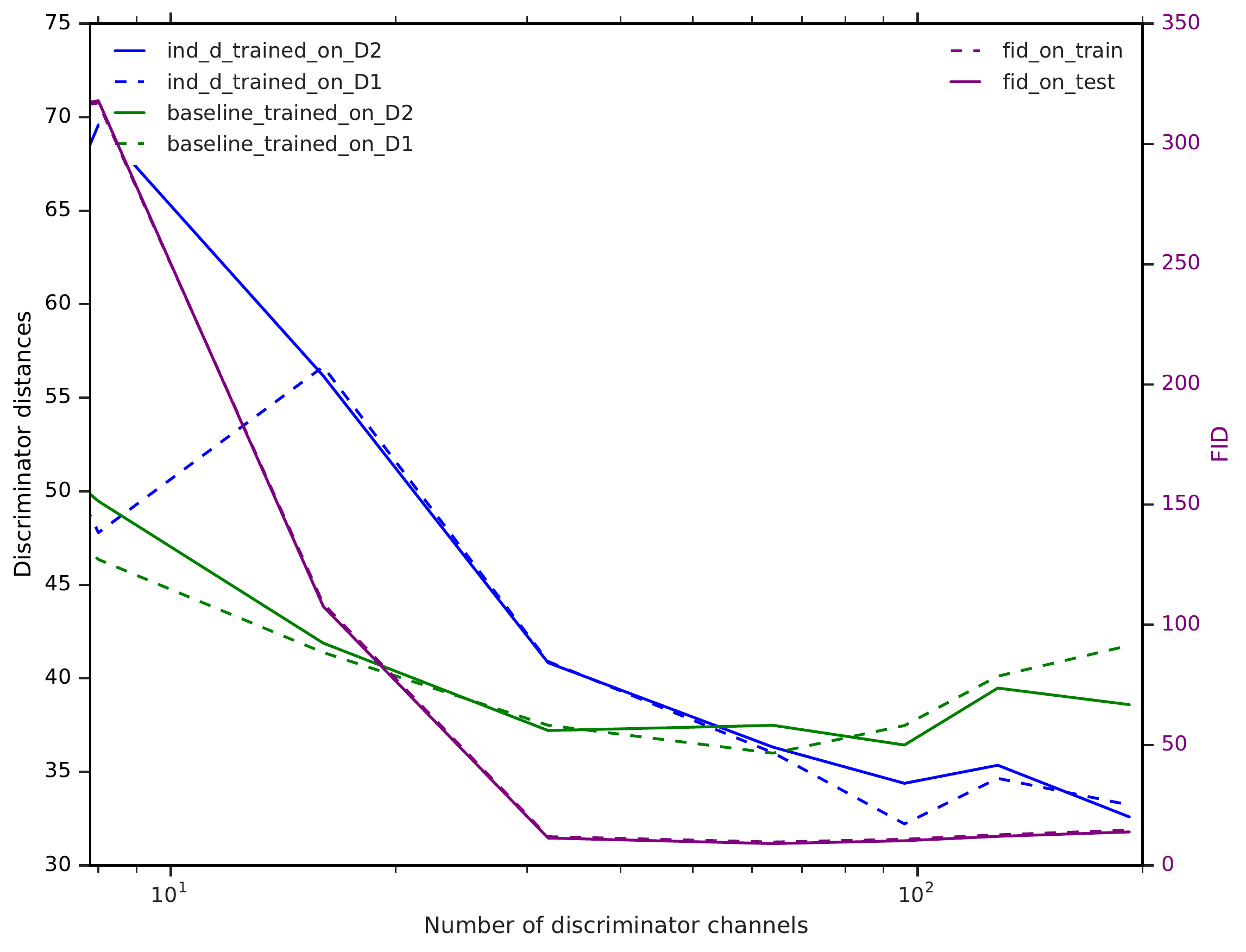}}
\end{center}
\vskip -0.2in
\end{figure}

%%%%%%%%%%%%%%%%%%%%%%%%%%%%%%%%%%%%%%%%%%%%%%%%%%%%%%%%%%%%%%%%%%%%%%%%%%%%%%%
%%%%%%%%%%%%%%%%%%%%%%%%%%%%%%%%%%%%%%%%%%%%%%%%%%%%%%%%%%%%%%%%%%%%%%%%%%%%%%%

\end{document}